\pdfoutput=1

\documentclass[11pt]{article}

\usepackage[]{EMNLP2023}

\usepackage{times}
\usepackage{latexsym}

\usepackage[T1]{fontenc}

\usepackage[utf8]{inputenc}

\usepackage{microtype}

\usepackage{inconsolata}

\usepackage{graphicx}
\usepackage{hyperref}
\usepackage{listings}
\usepackage{tabularx}
\usepackage{multirow}
\usepackage{easy-todo}
\usepackage{mathrsfs}
\usepackage[flushleft]{threeparttable}
\usepackage{enumitem}
\usepackage{amssymb}
\usepackage{tcolorbox}
\usepackage{fancyvrb}
\usepackage{xcolor}
\usepackage{etoolbox}

\newcommand{\ver}[1]{\texttt{#1}}

\newcommand{\tks}[1]{\rm\scriptscriptstyle{#1}}

\newtcolorbox{wordexample1}[1]
{
coltext=black,coltitle=black,colback=white,colbacktitle=white,colframe=black,
boxrule=1.0pt,
sharp corners,
left=2pt,right=2pt,top=2pt,bottom=2pt,
title={\begin{center}{\footnotesize #1}\end{center}}
}

\newtcolorbox{wordexample2}
{
coltext=black,coltitle=black,colback=white,colbacktitle=white,colframe=black,
boxrule=1.0pt,
sharp corners,
left=2pt,right=2pt,top=2pt,bottom=2pt,
}

\AtBeginEnvironment{wordexample1}{\scriptsize}
\AtBeginEnvironment{wordexample2}{\scriptsize}

\title{Matching of Descriptive Labels to Glossary Descriptions}

\author{Toshihiro Takahashi \\
	IBM Research - Tokyo \\
	\texttt{e30137@jp.ibm.com} \\\And
	Takaaki Tateishi \\
	IBM Research - Tokyo \\
	\texttt{tate@jp.ibm.com} \\\And
	Michiaki Tatsubori \\
	IBM Research - Tokyo \\
	\texttt{mich@jp.ibm.com}  \\
}

\begin{document}
\maketitle

\begin{abstract}

  Semantic text similarity plays an important role in software engineering tasks in which engineers are requested to clarify the semantics of descriptive labels (e.g., business terms, table column names) that are often consists of too short or too generic words and appears in their IT systems.
  We formulate this type of problem as a task of matching descriptive labels to glossary descriptions.
  We then propose a framework to leverage an existing semantic text similarity measurement (STS) and augment it using semantic label enrichment and set-based collective contextualization where the former is a method to retrieve sentences relevant to a given label and the latter is a method to compute similarity between two contexts each of which is derived from a set of texts (e.g., column names in the same table).
  We performed an experiment on two datasets derived from publicly available data sources.
  The result indicated that the proposed methods helped the underlying STS correctly match more descriptive labels with the descriptions.
\end{abstract}

\section{Introduction}

In general IT projects, such as database and business process migration, IT engineers invest significant effort in verifying consistency between various models, such as table schemata and diagrams that depict object relations.
Maintaining a glossary of domain terms is a best practice that helps alleviate their workload. The glossary serves as a reference that maps descriptive labels (such as business terms and table column names) to corresponding glossary descriptions, which are typically English sentences that provide explanations and contexts.

\subsection*{Target Problem and Challenges}
%
%
In this paper, we address such a common mapping problem between descriptive labels and glossary descriptions. We refer to this problem as Descriptive Labels to Descriptions (DLD). In DLD, we are given multiple datasets and glossaries. Each dataset contains a list of descriptive labels, while each glossary contains a list of glossary descriptions. The goal is to establish mappings from each label to its corresponding description.

%
%
However, there are several technical challenges in mapping descriptive labels to the glossary descriptions due to the nature of the descriptive labels such as too short and too generic words included by the descriptive names.

\subsection*{Our Approach and Research Questions}
To tackle the technical challenges, in this paper, we propose a novel framework to solve DLD problems effectively by enriching the short and/or generic words and capturing the context of the generic and/or ambiguous words.
Our framework is designed to be flexible enough to employ variations of underlying semantic text similarity (STS) models, enrichment methods, and contextualization methods whereas our implementation is limited to  reasonable combinations of a traditional TFIDF model, the PromCSE model~\cite{PromCSE2022} (BERT-based STS model), the Flan-T5~\cite{FlanT5} large language model (LLM), and Wikidata.
With our implementation we performed an experiment on two glossaries:
a business glossary derived from the financial industry business ontology and more than 1000 pairs of column names and corresponding descriptions obtained from the Kaggle webpages.
We organized our experiment to verify our hypothesis that there are many descriptive labels that cannot be mapped to corresponding descriptions by commonly used text similarity models due to cryptic words and our label enrichment and contextualization methods help the underlying STS models work better for such problematic labels.
More specially, the research questions we address in this paper are as follows.
\begin{itemize}
  \item How much improvement is observed in the metrics Mean Reciprocal Rank (MRR) and Hits@k with the label enrichment methods?
  \item How much improvement is observed in MRR and Hits@k with the label contextualization methods?
  \item What kind of labels meet the out-of-vocabulary issue and the ambiguity issue? And what kind of labels can be solved by the label enrichment methods, and can be disambiguated by label contextualization methods?
\end{itemize}
The reason why we use the Mean Reciprocal Rank (MRR) and Hits@k is that the DLD problem can be seen as a task of ranking descriptions corresponding to given descriptive labels and these metrics are commonly used for
recommender systems.

\subsection*{Contributions}

Our contributions in this paper are as follows.
\begin{itemize}[itemsep=0pt,parsep=0pt]
  \item We formulate the DLD problem, and identified the technical issues: out-of-vocabulary issue and ambiguity issue of cryptic words. We provide two practical benchmark datasets: Kaggle and FIBO including these issues.
  \item To solve out-of-vocabulary issue, we propose Label Semantic Enrichment method which leverages external knowledge.
  \item To solve ambiguity issue, we propos Set-based Collective Contextualization method by leveraging Large Language Model.
  \item In our experiments, we clarified how effectively our approaches solve the issues.
\end{itemize}

The rest of this paper is organized as follows.
In the next two sections, we discuss related works, and provide motivating examples using practical datasets.
We then present our approach to address them effectively, and formulate the DLD task.
Subsequently, we demonstrate how the DLD task is tackled using several approaches, presenting experimental results. Afterward, we provide a comprehensive analysis of our experimental results.
Finally, we conclude the paper.

\section{Related Work}

In general, DLD is a task of linking a chunk of text in a group to one in another group.
In this sense, it could be considered as a natural language processing problem, specifically semantic text similarity or entity linking~\cite{EntityLinking, Shen+:2021:EntityLinkingDL}.  Also, we could consider it as a problem of aligning between domain-specific labels or descriptive names, as a set of domain concepts, to another ontology of a common glossary.

\subsection*{Semantic Text Similarity}

There have been many studies on STS~\cite{chan2021evolution}.
Some exploited general knowledge bases~\cite{li2020efficient}, as in our study.
Such knowledge-based methods measure the similarity of two terms on the basis of the
structural properties of the knowledge bases, such as the number of edges.
Our method, however, uses the knowledge bases only to enrich descriptive names.

Other methods are corpus based such as word2vec~\cite{word2vec} and BERT~\cite{BERT}.
Such methods leverage large corpora to compute word-embeddings useful
for measuring the similarity between terms on the basis of the idea that similar words
occur together.
The same idea is also applied for capturing the characteristics of sentences,
as with Sentence-BERT~\cite{SBERT2019} and PromCSE~\cite{PromCSE2022}.
Many search engines use this type of method to retrieve and rank relevant
sentences and webpages.
We designed our method in such a way that it benefits from the advances of
the corpus-based methods and pre-trained models.

\subsection*{Entity Linking and Ontology Learning}

Entity linking~\cite{EntityLinking}, such as ColNet~\cite{ColNet2019} and TabEL~\cite{TabEL2015},
is a task of linking terms described in a document to entities defined in a knowledge
graph such as Wikidata.
Ontology learning~\cite{asim2018survey}, matching~\cite{shvaiko2011ontology}, or alignment~\cite{Ardjani+:2015:OntologyAlignment}
are similar tasks that automatically or semi-automatically gather terms and
relations from documents to create an ontology and discover correspondence between ontologies.
Background knowledge is known to significantly improve the performance of ontology matching systems~\cite{Portisch+:ISWC2021:BKSM}.
Compared with these problems, the DLD problem mentions neither knowledge graph
nor ontologies.
The descriptive names in a DLD problem are not within contextual sentences or linked to
other concepts, entities, or values.
Glossaries as ontologies in a DLD problem are also not linked but just descriptions of known concepts.

\subsection*{Named Entity Disambiguation}

Named entity disambiguation on knowledge graphs including ontologies is a key component for the success of semantic text similarity, entity linking, and ontology learning.
Likewise, our label enrichment method and set-based collective contextualization are both considered as methods to disambiguate the descriptive labels for solving the DLD problems.
As in the case of our label enrichment method, recent literatures~\cite{mulang2020,bastos2021recon} also leverage the triples obtained from Wikidata to improve the performance of pre-trained models for the named entity disambiguation on Wikipedia whereas we leverage the sentences obtained from Wikidata.
In addition, unlike our contextualization, they do not compute the context of a set of entities and rely only on the context of each single entity.

The idea of using a knowledge base for disambiguation is also presented in the literature~\cite{ho2010word}.
It proposes the use of a knowledge base to measure the similarity between texts for semantic text similarity and integrate it with a corpus-based measurement.
However, it does not include any process for enriching and/or contextualizing given texts.

\section{Motivating Examples}
\label{sec:motivating}

In software engineering tasks such as database and business process migrations, engineers are often requested to clarify the semantics of descriptive labels (e.g., business terms, table column names).
For example, during the migration from a legacy system, IT engineers thoroughly analyze the original descriptive labels present in the existing data models.
They then create a new logical data model and establish mappings between the original labels and the new ones.
Understanding the semantics of the original labels in the context of the new logical data model is crucial, and the glossary serves as a valuable resource.
However, the original data model often suffers from incompleteness and inconsistencies, as its element names may differ from those in the new logical data model, and an up-to-date or accessible dictionary may be lacking. Moreover, the mapping process typically requires human involvement and may not have an initial mapping in the first iteration.

We could leverage a state-of-the-art semantic text similarity (STS) measurement to map each descriptive labels to a corresponding description. However, the following nature of the descriptive names makes it difficult to make the mapping using the STS measurement in high accuracy.
\begin{itemize}
  \item The descriptive labels (e.g., LOAN\_AMT, ACS\_DT) often contain cryptic (too short, too generic, and too ambiguous) words such as AMT (amount), ACS (access), and DT (date). These cryptic words are out-of vocabulary unlike aliases or nicknames, and often makes it difficult to understand their meanings.
  \item The descriptive labels often appear only in database table schemata or program variables. This prevents us from associating the descriptive labels with documents.
  \item Many descriptive labels used in the IT systems are specific to those IT systems. We could not rely on mappings between the descriptive labels and descriptions created for other IT systems.
\end{itemize}

The same situation arises even in standardized or commonly used domain-specific ontologies such as financial industry business ontology (FIBO).
FIBO is an ontology for financial business applications.
It defines a named entity ``ALL'' the description of which is
``the currency identifier for Lek (the currency of Albania)''.
However, it is a very general word and often used as in the case of ``all types of bank loans''.
Likewise, we found many descriptions that include the word ``all''.
Therefore, no STS model is effective for
matching the descriptive name with the correct definition.

As a preliminary experiment, we collected the named entities and corresponding descriptions from FIBO,
and ran PromCSE to compute the similarity scores between all the pairs
of the named entities and descriptions.
The top-ranked descriptions corresponding to ``ALL'' were
\begin{itemize}[itemsep=0pt,parsep=0pt]
  \item ``collection representing the total membership, or \'universe\', of people, resources, products, services, events, or entities of interest for some question, experiment, survey or statistical program'' (0.309),
  \item ``location in physical space'' (0.291),
  \item ``a collection of managed investments that are all managed by a single investment institution'' (0.274),
\end{itemize}
where the values enclosed with the parentheses are the similarity scores reported from PromCSE.
The correct description of ``ALL'' was ranked 490th (0.063).

\section{Approach}
\label{sec:approach}
Our approach is as follows:
\begin{itemize}[itemsep=0pt,parsep=0pt]
  \item We employ a STS model to measure a similarity of two sentences.
  \item We apply Label Semantics Enrichment module to enrich descriptive labels.
  \item We take into account the context of both of descriptive labels and glossary descriptions.
\end{itemize}

\subsection*{Label Semantics Enrichment (LSE)}
A basic strategy to measure a similarity between a descriptive label and a glossary description is using STS models such as TFIDF, PromCSE and LLM.
TFIDF suffer from aliases because it's built on word-level exact matching.
PromCSE and LLM also may suffer from minor aliases and cryptic words in descriptive labels.

Furthermore we employ Label Semantics Enrichment (LSE) to solve this problem.
LSE module retrieves sentences relevant to the given the descriptive label by using a external knowledge database such as Wikidata and Bing.
If the retrieved sentences include sufficiently various aliases and relevant phrases of cryptic words, STS model will work more effectively.

\subsection*{Set-based Collective Contextualization (SCC)}
As mentioned earlier, descriptive labels can often be ambiguous. In the case that two tables have columns with the same label but semantically different meanings, the same glossary description will be assigned to columns that are semantically different.

To address this problem, we propose incorporating the context of a set of descriptive labels and a set of glossary descriptions.
For example, when matching column names from multiple tables to various glossaries, we consider the collective context provided by a set of column names within the same table and a set of glossary descriptions within the same glossary.
By leveraging this broader context, we can more effectively identify and disambiguate the intended meanings of the columns.



\section{Problem Setting}
\label{sec:problem_setting}

Now we are given a dataset ${\mathcal D}$ which contains several semantic groups $D_i \in {\mathcal D}$.
The i-th group $D_i = (L_i, G_i)$ has a set of descriptive labels $L_i = \{l_{i,1}, l_{i,2}, \cdots, l_{i,n}\}$ and a glossary (a set of glossary descriptions) $G_i = \{g_{i,1}, g_{i,2}, \cdots, g_{i,n}\}$.
$l_{i,p} \in L_i$ is p-th descriptive label of $L_i$.
$g_{i,p} \in G_i$ is p-th glossary description of $G_i$.
$g_{i,p}$ describes a meaning of $l_{i,p}$.

The ultimate goal is to establish a complete mapping between $(l_{i,p}, L_i)$ and its corresponding $(g_{i,p}, G_i)$. However, the difficulty of this problem is heavily influenced by the number of groups $D_i$ and the size of the label set $L$ and the glossary $G$ in each group $D$.

To standardize the difficulty, we consider N-choice problem here. In this scenario, given a target label set $L$ and a target label $l \in L$, the objective is to identify the corresponding pair of the target description and its glossary $(g, G)$ from N candidates $\{(g_k, G_k)\}_{k=1}^N$. This is done by measuring the similarity between the label side $(l, L)$ and the glossary side $(g_k, G_k)$.


\section{DLD Framework}

In our framework, DLD-Similarity Score $\Psi$ between label side $(l, L)$ and glossary side $(g, G)$ is defined as
\begin{eqnarray*}
  \Psi(l, g, L, G | \theta) &=& \Psi_{\rm T}(l, g, L | \theta_{\tks LSE}, \theta_{\tks STS}) \\
  & \times& \Psi_{\rm C}(L, G | \theta_{\tks SCC}).
\end{eqnarray*}
Here $\Psi_{\rm T}$ is Text-Similarity Score, and $\Psi_{\rm C}$ is Context-Similarity Score.
In $\Psi_{\rm T}$, $\theta_{\tks LSE} \in \{ {\rm on}, {\rm off} \}$ is a switch to enable or disable LSE module, and $\theta_{\tks STS} \in \{ \ver{T}, \ver{P}, \ver{L} \}$ is a switch indicating which STS model is used from three variations: TFIDF, PromCSE, and LLM.
In $\Psi_{\rm C}$, $\theta_{\tks SCC} \in \{ {\rm on}, {\rm off} \}$ is a switch to enable or disable SCC module.

\subsection*{Text-Similarity Score}
Text-Similarity Score $\Psi_{\rm T}$ measures a similarity between $l$ and $g$.
It's defined as
\begin{eqnarray*}
  \Psi_{\rm T}(l, g, L | \theta_{\tks LSE}, \theta_{\tks STS}) = \\
  \max_{s \in {\rm LSE}(l | \theta_{\tks LSE})} {\rm STS}(s, g, L | \theta_{\tks STS}).
\end{eqnarray*}
$\Psi_{\rm T}$ collects relevant sentences $\{s_i\}$ by invoking ${\rm LSE}$, and computes similarity score between each sentence $s_i$ and $g$, and outputs max of them.

We have two variations of ${\rm LSE}$ function: enabled version and disabled version.
The enabled version ${\rm LSE}(l | {\rm on})$ collects sentences $\{s_i\}$ relevant to $l$ from external knowledge, and returns them.
The disabled version ${\rm LSE}(l | {\rm off})$ just returns $l$.

We also have three variations of ${\rm STS}$ function: TFIDF version ${\rm STS}(\cdot | \ver{T})$, PromCSE version ${\rm STS}(\cdot | \ver{P})$, and LLM version ${\rm STS}(\cdot | \ver{L})$.
See \ref{sec:llm-based_sts} for more details.

\subsection*{Context-Similarity Score}
Context-Similarity Score $\Psi_{\rm C}$ measures a contextual similarity between $L$ and $G$.
We have two variations: SCC enabled version and disabled version.

The SCC enabled version $\Psi_{\rm C}(L, G | {\rm on})$ directly ask to LLM about a probability of $L$ and $G$ being the same or different.
See \ref{ssec:llm-based_context-similarity} for more details.

The disable version $\Psi_{\rm C}(L, G | {\rm off})$ is always outputs $1$.


\section{Implementation Details}

We implemented our method using Python.
It leverages Wikidata as external knowledge in LSE module.
It runs TFIDF, PromCSE and Flan-T5~\cite{FlanT5} in STS module to measure a similarity score between two sentences, and also uses Flan-T5 in SCC module to measure Context-Similarity Score between a set of labels and a set of descriptions.
The use of TFIDF and PromCSE is straightforward,
therefore, in the following sections, we describe about how to use Wikidata in LSE module, and how to use Flag-T5 in STS and SCC module.

\subsection*{Using Wikidata for LSE}
\label{sec:using_wikidata_for_lse}

In LSE module, we used Wikidata as external resource.
The webpage of Wikidata provides a search interface that we usually access using a Web browser.
We leverage this Web interface for the implementation.
It sends queries of descriptive labels to the search interface using the HTTP protocol, and parses resulting webpages to extract entity IDs.

Wikidata also provides a SPARQL~\citep{SPARQL} endpoint as a query service.
We leverage this query service to collect label names and descriptions of the collected entity IDs.
We also use the \texttt{label} property\footnote{http://www.w3.org/2000/01/rdf-schema\#label} and \texttt{description} property\footnote{https://schema.org/description} to generate sentences relevant to the query phrase.

\subsection*{LLM-based Semantic Text Similarity Model}
\label{sec:llm-based_sts}

We employed three STS models (TFIDF, PromCSE, and LLM) to compute the Text-Similarity Score $\Psi_{\rm T}$.
TFIDF and PromCSE can directly generate similarity scores for the given pair of sentences. On the other hand, LLM requires a natural language prompt as input, and also produces a natural language answer.

We performed prompt engineering to create suitable inputs for LLM and developed a method to extract scores from the generated answer.
Figure \ref{fig:prompt_example_for_llm_sts} shows the typical prompt example.

To translate the LLN answer to numerical score, ${\rm STS}(\cdot | \ver{L})$ function collects top N tokens ${\mathscr T} = \{t_i\}_{i=1}^N$ with their probability $p(t_i)$ from LLM answer.
It classifies the N tokens into a set of ``yes'' tokens ${\mathscr T}_{\rm y}$, ``no'' tokens ${\mathscr T}_{\rm n}$, and others.
And computes probability ratio of ``yes'' and ``no'' as $s = p_{\rm y}/(p_{\rm y} + p_{\rm n})$.
Here $p_{\rm y}$ and $p_{\rm n}$ is sum of probability which token is ``yes'' and ``no''.
They can be computed as follows:
$$
  p_{\rm y} = \sum_{t \in {\mathscr T}_{\rm y}} p(t).
$$

\begin{figure*}[tbh]
  \begin{lstlisting}[escapechar=\!]
I have a dataset and a glossary. The given dataset has these columns.
[Column names]
  - School District Code
  - County Code
  ...
[Question]
Is "School District Code" same to the following concept in glossary?
glossary description: "The code by which a school district is identified, as utilized by the Department!’!s ..."
[Answer (Yes/No)]
\end{lstlisting}
  \vspace{-1em}
  \caption{Prompt example for LLM-based STS}
  \label{fig:prompt_example_for_llm_sts}
\end{figure*}

\subsection*{LLM-based Context-Similarity Algorithm}
\label{ssec:llm-based_context-similarity}

We also employed LLM to measure the Context-Similarity between a set of labels and a set of descriptions.
Figure \ref{fig:prompt_example_for_scc} shows the typical prompt example.
A scoring logic is same to the logic mentioned in \ref{sec:llm-based_sts}.

\begin{figure*}[tbh]
  \begin{lstlisting}[escapechar=\!]
I have a dataset and a glossary. The given dataset has these columns.
[Column names]
  - ordered
  - device_computer
  ...
The given glossary has these glossary terms.
[Glossary terms]
  - Indicates whether an appeal to the published decision has been received.
  - Indicates if Design Review is part of the application process for this permit.
  ...
[Question]
Does these glossary terms describe the given column names? 
[Answer (Yes/No)]
\end{lstlisting}
  \vspace{-1em}
  \caption{Prompt example for SCC}
  \label{fig:prompt_example_for_scc}
\end{figure*}







\section{Experimental Setup}
\label{sec:experimental_setup}

As we described in Section 1, our experiment is organized to verify the hypothesis that (1) there are many descriptive labels that cannot be mapped to corresponding descriptions by commonly used text similarity models due to cryptic words and (2) our label enrichment and contextualization methods help the underlying STS models work better for such problematic labels.
In this section, we first describe how we prepared the datasets based on the publicly available data sources.
We then describe what metrics and why we used and how we compared the different combinations of the STS models, the label enrichment method, and the contextualization method.



\subsection*{Dataset and Benchmark}
\label{ssec:dataset_and_benchmark}

Our experiment is performed on the two datasets derived from the Kaggle webpages and the financial industry business ontology (FIBO).
Table \ref{tbl:statistics} shows the statistics of these datasets.

The Kaggle dataset consists of 85 semantic groups that includes 1347 descriptive labels and corresponding descriptions in total where we consider tables and column names we extracted from the Kaggle webpages as the semantic groups and the descriptive labels.
The smallest semantic group contains only 10 descriptive labels while the largest semantic group contains 36 descriptive labels.
We further investigated how many semantic groups include each descriptive label to see if the dataset is suitable to evaluate the effectiveness of the contextualization method.
Table \ref{tbl:freq_column_name} summarizes the descriptive labels and the numbers of corresponding semantic groups, which are referred to as frequencies of the descriptive labels, where we selected only the descriptive labels whose frequency is more than 7.
For example, the descriptive label ``type'' appears in 7 semantic groups and has different meanings such as
``type of wine'', ``media type of animation film'' and ``flag if company is private or public''.

FIBO defines concepts and relations used in financial domain, using the Web Ontology Language (OWL)~\citep{OWL}.
It consists of 2086 named entities each of which has its label and description specified by particular XML tags such as the \texttt{description} and \texttt{definition}.
We collected these labels and descriptions as descriptive labels and descriptions of the dataset.
We then partitioned the pairs of the descriptive labels and descriptions into 44 semantic groups based on IRIs of the corresponding named entities.
Unlike the Kaggle dataset, there is no descriptive label that is included by multiple semantic groups.

From these two datasets, we created 4 problems: Kaggle-10-choice, Kaggle-50-choice, FIBO-10-choice, FIBO-50-choice as mentioned in Section \ref{sec:problem_setting} and performed the experiment on these 4 problems.

\begin{table}[tbh]
  \begin{center}
    \caption{Statistics of the datasets}
    \label{tbl:statistics}
    \begin{footnotesize}
      \begin{tabular}{l|rr}
        \hline
                                   & Kaggle & FIBO \\
        \hline
        entries                    & 1347   & 2086 \\
        semantic groups            & 85     & 44   \\
        avg of \# of words (label) & 1.64   & 4.05 \\
        max of \# of words (label) & 6      & 16   \\
        min of \# of words (label) & 1      & 1    \\
        avg of \# of words (desc)  & 14.4   & 16.1 \\
        max of \# of words (desc)  & 389    & 140  \\
        min of \# of words (desc)  & 2      & 3    \\
        \hline
      \end{tabular}
    \end{footnotesize}
  \end{center}
\end{table}

\begin{table}[tbh]
  \begin{center}
    \caption{Frequency of descriptive labels of Kaggle}
    \label{tbl:freq_column_name}
    \begin{footnotesize}
      \begin{tabular}{lr|lr}
        \hline
        label  & frequency & label       & frequency \\
        \hline
        name   & 16        & country     & 8         \\
        date   & 14        & title       & 8         \\
        age    & 12        & id          & 8         \\
        year   & 9         & type        & 7         \\
        county & 9         & status      & 7         \\
        gender & 8         & description & 7         \\
        \hline
      \end{tabular}
    \end{footnotesize}
  \end{center}
\end{table}


\subsection*{Evaluation Method}
\label{ssec:methods_compared}

%

In our evaluation, we compare the 12 combinations of models based on the 3 underlying STS models (TFIDF, PromCSE, LLM-based) and the presence or absence of the label semantics enrichment (SLE) and the set-based collective contextualization (SCC), which are represented by the following naming rule:
$
  \{\ver{T}, \ver{P}, \ver{L}\}{\mathchar`-}\{\phi, \ver{LSE}\}{\mathchar`-}\{\phi, \ver{SCC}\},
$
where $\ver{T}$,$\ver{P}$, and $\ver{L}$ represent TFIDF, PromCSE, and the LLM-based STS.
For example, `\ver{T-LSE-SCC}' represents the combination of TFIDF, LSE, and SCC.
`\ver{L}' represents the LLM-based STS model not augmented with LSE or SCC.

To measure the success of matching the descriptive labels with the descriptions, we employ the Mean Reciprocal Rank (MRR) and Hits@k as performance metrics which are commonly used for evaluating the performance of search engines and recommendation systems.
This is because the DLD problem can be seen as a task of ranking descriptions corresponding to given descriptive labels.


\section{Experimental Results}

We present our comprehensive results against the 10-choice problems of the Kaggle and FIBO datasets in Table \ref{tbl:ex_result_of_10c} and against the 50-choice problems in Table \ref{tbl:ex_result_of_50c}.

As a whole, regarding the first and second research questions, we proved that the label semantics enrichment (LSE) and the set-based collective contextualization (SCC) helps the underlying STS models produce better scores in both MRR and Hits@k except that LSE often gave negative impact to PromCSE and the LLM-based STS model in many cases.

Regarding the third research question, we observed that there were reasonably many labels not correctly mapped only by the underlying STS models were correctly mapped by the STS model augmented with LSE and SCC. In particular, SCC successfully disambiguated generic words, and helped the underlying STS model correctly mapped out-of-vocabulary labels which include highly cryptic words.
In addition, the \ver{L-SCC} combination achieved over 99\% in MRR for the 10-choice problem. The failed descriptive labels were difficult to be correctly mapped even by humans.

We describe more details in the following sections one by one.


\begin{table}[tbh]
  \caption{Result of 10-choice problem}
  \label{tbl:ex_result_of_10c}
  \vspace{-1em}
  \begin{center}
    \begin{footnotesize}
      \begin{tabular}{l|cccc}
        \hline
        Kaggle10C       & MRR            & Hits@1         & Hits@3         & Hits@5         \\
        \hline
        \ver{T}         & 0.735          & 0.670          & 0.687          & 0.687          \\
        \ver{T-LSE}     & 0.795          & 0.733          & 0.788          & 0.818          \\
        \ver{T-SCC}     & 0.741          & 0.682          & 0.687          & 0.687          \\
        \ver{T-LSE-SCC} & 0.864          & 0.827          & 0.854          & 0.860          \\
        \ver{P}         & 0.872          & 0.800          & 0.931          & 0.972          \\
        \ver{P-LSE}     & 0.865          & 0.794          & 0.918          & 0.963          \\
        \ver{P-SCC}     & 0.987          & 0.978          & 0.996          & 0.999          \\
        \ver{P-LSE-SCC} & 0.986          & 0.977          & 0.996          & 0.997          \\
        \ver{L}         & 0.982          & 0.970          & 0.995          & 0.997          \\
        \ver{L-LSE}     & 0.975          & 0.958          & 0.993          & 0.997          \\
        \ver{L-SCC}     & 0.994          & \textbf{0.990} & \textbf{0.998} & \textbf{0.999} \\
        \ver{L-LSE-SCC} & \textbf{0.994} & \textbf{0.990} & \textbf{0.998} & \textbf{0.999} \\
        \hline
        FIBO10C         & MRR            & Hits@1         & Hits@3         & Hits@5         \\
        \hline
        \ver{T}         & 0.751          & 0.689          & 0.716          & 0.727          \\
        \ver{T-LSE}     & 0.874          & 0.833          & 0.881          & 0.907          \\
        \ver{T-SCC}     & 0.768          & 0.712          & 0.729          & 0.733          \\
        \ver{T-LSE-SCC} & 0.925          & 0.898          & 0.933          & 0.937          \\
        \ver{P}         & 0.941          & 0.911          & 0.963          & 0.986          \\
        \ver{P-LSE}     & 0.957          & 0.934          & 0.973          & 0.990          \\
        \ver{P-SCC}     & 0.988          & 0.978          & 0.999          & \textbf{1.000} \\
        \ver{P-LSE-SCC} & 0.990          & 0.981          & \textbf{1.000} & \textbf{1.000} \\
        \ver{L}         & 0.988          & 0.978          & 0.998          & \textbf{1.000} \\
        \ver{L-SCC}     & \textbf{0.993} & \textbf{0.986} & \textbf{1.000} & \textbf{1.000} \\
        \hline
      \end{tabular}
    \end{footnotesize}
  \end{center}
  \begin{tablenotes}
    \begin{footnotesize}
      \item This table shows the experiment results under the 12 settings (STS (TFIDF, PromCSE, LLM), LSE on/off, SCC on/off) with Kaggle and FIBO datasets.
      Hit@k is hit ratio represents how many times the true descriptions appeared in top-$k$ in the list. Larger is better.
      MRR is mean reciprocal rank. Larger is better.
      Highest scores of each metric are denoted in \textbf{bold}.
    \end{footnotesize}
  \end{tablenotes}
\end{table}

\begin{table}[tbh]
  \caption{Result of 50-choice problem}
  \label{tbl:ex_result_of_50c}
  \vspace{-1em}
  \begin{center}
    \begin{footnotesize}
      \begin{tabular}{l|cccc}
        \hline
        Kaggle50C       & MRR            & Hits@1         & Hits@5         & Hits@10        \\
        \hline
        \ver{T}         & 0.654          & 0.604          & 0.686          & 0.687          \\
        \ver{T-SLE}     & 0.698          & 0.635          & 0.753          & 0.776          \\
        \ver{T-SCC}     & 0.691          & 0.671          & 0.687          & 0.687          \\
        \ver{T-SLE-SCC} & 0.816          & 0.782          & 0.842          & 0.849          \\
        \ver{P}         & 0.726          & 0.624          & 0.846          & 0.916          \\
        \ver{P-SLE}     & 0.728          & 0.638          & 0.836          & 0.906          \\
        \ver{P-SCC}     & 0.960          & 0.933          & 0.990          & 0.995          \\
        \ver{P-SLE-SCC} & 0.957          & 0.928          & 0.989          & 0.993          \\
        \ver{L}         & 0.939          & 0.903          & 0.984          & 0.993          \\
        \ver{L-SCC}     & \textbf{0.982} & \textbf{0.970} & \textbf{0.993} & \textbf{0.996} \\
        \hline
        FIBO50C         & MRR            & Hits@1         & Hits@5         & Hits@10        \\
        \hline
        \ver{T}         & 0.683          & 0.648          & 0.701          & 0.712          \\
        \ver{T-SLE}     & 0.810          & 0.772          & 0.848          & 0.868          \\
        \ver{T-SCC}     & 0.707          & 0.677          & 0.720          & 0.728          \\
        \ver{T-SLE-SCC} & 0.877          & 0.841          & 0.916          & 0.928          \\
        \ver{P}         & 0.870          & 0.824          & 0.923          & 0.952          \\
        \ver{P-SLE}     & 0.907          & 0.871          & 0.950          & 0.969          \\
        \ver{P-SCC}     & 0.953          & 0.918          & 0.995          & 0.998          \\
        \ver{P-SLE-SCC} & 0.965          & 0.938          & 0.998          & \textbf{1.000} \\
        \ver{L}         & 0.958          & 0.932          & 0.992          & 0.996          \\
        \ver{L-SCC}     & \textbf{0.976} & \textbf{0.957} & \textbf{0.999} & \textbf{1.000} \\
        \hline
      \end{tabular}
    \end{footnotesize}
  \end{center}
\end{table}

\subsection*{Effectiveness of LSE}

We observed that LSE was effective only for TFIDF models as shown in Table \ref{tbl:improvement_by_lse_in_kaggle10}, but the negative impact to the other models are very limited.
We think that the pre-trained models, PromCSE and LLM, already have better or equivalent capability compared with LSE.

Table \ref{tbl:general_by_lse} shows some examples of descriptive labels whose rankings were improved or degraded by LSE from the results of Kaggle-10-choice problem.
It shows that LSE properly address the out-of-vocabulary problem encountered in the TFIDF model, thereby successfully enriched cryptic or domain specific words.
Additionally, we provide some examples of descriptive labels from the Kaggle result that were accurately ranked as top 1 by the LSE-enhanced model (\ver{T-LSE}), but failed to achieve the same ranking by the baseline model (\ver{T}) in Table \ref{tbl:general_by_lse}.

On the other hand, there were certain descriptive labels that were successfully ranked by the baseline model (\ver{T}), but failed by the LSE-enhanced model (\ver{T-LSE}).
Table \ref{tbl:general_by_lse} shows some examples.
For more general words, the incorporation of LSE may introduce noise and hinder the accurate identification of the corresponding glossary description.

\begin{table}[tbh]
  \caption{Improved MRR by LSE in Kaggle10}
  \label{tbl:improvement_by_lse_in_kaggle10}
  \vspace{-1em}
  \begin{center}
    \begin{footnotesize}
      \begin{tabular}{lc|lc|c}
        \hline
        w/o LSE     & MRR   & w/ LSE          & MRR   & improved       \\
        \hline
        \ver{T}     & 0.735 & \ver{T-LSE}     & 0.795 & 0.061          \\
        \ver{T-SCC} & 0.741 & \ver{T-LSE-SCC} & 0.864 & \textbf{0.123} \\
        \ver{P}     & 0.872 & \ver{P-LSE}     & 0.865 & -0.007         \\
        \ver{P-SCC} & 0.987 & \ver{P-LSE-SCC} & 0.986 & -0.001         \\
        \ver{L}     & 0.982 & \ver{L-LSE}     & 0.975 & -0.007         \\
        \ver{L-SCC} & 0.994 & \ver{L-LSE-SCC} & 0.994 & 0.000          \\
        \hline
      \end{tabular}
    \end{footnotesize}
  \end{center}
\end{table}

\begin{table}[tbh]
  \caption{Example descriptive labels effected by LSE}
  \label{tbl:general_by_lse}
  \vspace{-1em}
  \begin{center}
    \begin{wordexample1}{Improved labels by LSE}
      `AADT', `Burglary', `chlorides', `DEROG', `FGA', `FTA', `HAZMAT', `isbn', `isFlaggedFraud', `iso3', `lvdd', `MentHlth', `MVP', `Riot', `rpm', `Sugarcanes', `synopsis', `total sulfur dioxide', `Ward Interactions'
    \end{wordexample1}
    \begin{wordexample1}{Degraded labels by LSE}
      `Age', `category', `Category', `country', `date', `description', `Genre', `id', `Make', `name', `Rating', `Score', `sex', `status', `Status', `Title', `type', `url'
    \end{wordexample1}
  \end{center}
\end{table}

\subsection*{Effectiveness of SCC}

Table \ref{tbl:improvement_by_scc_in_kaggle10} shows the improvement of the MRR scores by SCC
where SCC improved the MRR score in all the cases.
In particular, it contributed to the improvement of over 10\% for the PromCSE models.

We analyzed the descriptive labels that were successfully ranked as top 1 as the result of the improvement by SCC (\ver{P-LSE-SCC}), but failed by the baseline model (\ver{P-LSE}).
We found these descriptive labels can be classified into two groups: general words and highly cryptic words.
For general words, PromCSE collected several descriptions from different glossaries with high confidences. SCC worked as screening indicators in this situation.
Table \ref{tbl:general_by_scc} shows some examples of the general labels whose rankings were improved by SCC.

In the case of highly cryptic words, PromCSE often struggled to find similar descriptions from all candidates, because the highly cryptic words are out-of-vocabulary even for PromCSE.
Consequently, the Text-Similarity Scores of the top-ranked descriptions were relatively low.
However, the Context-Similarity Score of the corresponding glossary was significantly high. SCC elevated the confidence of these correct but low-ranked descriptions, pushing them to the top of the list.
Table \ref{tbl:general_by_scc} also shows some examples of the cryptic labels improved by SCC.

\begin{table}[tbh]
  \caption{Improved MRR by SCC in Kaggle10}
  \label{tbl:improvement_by_scc_in_kaggle10}
  \vspace{-1em}

  \begin{center}
    \begin{footnotesize}
      \begin{tabular}{lc|lc|c}
        \hline
        w/o SCC     & MRR   & w/ SCC          & MRR   & improved       \\
        \hline
        \ver{T}     & 0.735 & \ver{T-SCC}     & 0.741 & 0.006          \\
        \ver{T-LSE} & 0.795 & \ver{T-LSE-SCC} & 0.864 & 0.069          \\
        \ver{P}     & 0.872 & \ver{P-SCC}     & 0.987 & 0.115          \\
        \ver{P-LSE} & 0.865 & \ver{P-LSE-SCC} & 0.986 & \textbf{0.122} \\
        \ver{L}     & 0.982 & \ver{L-SCC}     & 0.994 & 0.011          \\
        \ver{L-LSE} & 0.975 & \ver{L-LSE-SCC} & 0.994 & 0.019          \\
        \hline
      \end{tabular}
    \end{footnotesize}
  \end{center}

\end{table}

\begin{table}[tbh]
  \caption{Example descriptive labels improved by SCC}
  \label{tbl:general_by_scc}
  \vspace{-1em}
  \begin{center}
    \begin{wordexample1}{General labels}
      `Age', `close', `Close', `date', `Gender', `high', `id', `ID', `name', `Name', `Pos', `Services', `Status', `Tags', `Title', `TITLE', `type', `Type', `url', `Value', `y', `Year', `Years'
    \end{wordexample1}
    \begin{wordexample1}{Cryptic labels}
      `3PA', `ACS/Map', `AST', `CholCheck', `CLAGE', `CLNO', `DEROG', `DREB', `Fedu', `FGS', `FTA', `FTM', `G1', `G2', `G3', `GD', `GF', `GO / SC Num', `hsc\_p', `MentHlth', `Mjob', `OREB', `PTS', `RC\_ID', `RH', `SFY'
    \end{wordexample1}
  \end{center}
\end{table}



\section{Concluding Remarks}
\label{sec:conclusion}

We formulated the DLD problem as important and practical task,
and identified the technical issues: out-of-vocabulary issue and ambiguity issue of cryptic words.
We proposed a framework to solve the issues.
To solve out-of-vocabulary issue, we proposed Label Semantic Enrichment method by leveraging external knowledge.
To solve ambiguity issue, we proposed Set-based Collective Contextualization method by leveraging Large Language Model.
We provided two practical benchmark datasets: Kaggle and FIBO including the issues,
and designed N-choice problem on the datasets.
In our experiments, we clarified how our approach are effective to solve the issues.
We plan to release the benchmark datasets under a reasonable license for future advances
in this technical area.


\section*{Limitations}

The empirical evaluation of our methods is mainly done on the datasets derived from
the publicly available data sources whereas we used pre-trained models of Flan-T5 and PromCSE in the evaluation.
Therefore, there might be overlapping data sources, and hence the risk of data leakage.
Even so, our evaluation showed that both the label enrichment and the contextualization contributed to the improvement of the TFIDF-based STS, which never rely on any external data sources.

In our experiment, the LLM-based STS model outperformed the other models in all the cases.
However, we need to care about its inference time when we use it in a practical situation, since the estimated total inference time for completing the experiment was roughly 150 hours.
On the other hand, the TFIDF and PromCSE models were obviously more efficient than the LLM-based STS model. Those total inference times based on our observation were 1.6 hours and 25 hours, respectively.


\bibliography{main}
\bibliographystyle{acl_natbib}



\end{document}